\documentclass[letterpaper, 10 pt, journal, twoside]{ieeetran}  

\usepackage{graphicx} 
\usepackage{tensor}
\usepackage[font=footnotesize]{caption}
\usepackage{cite}
\usepackage[font=normal]{subcaption}
\usepackage{booktabs}
\usepackage{times} 
\usepackage{bm}
\usepackage{amsmath,amssymb,mathrsfs,mathdesign,upgreek,dsfont,gensymb}
\usepackage[ruled,vlined]{algorithm2e}
\usepackage{verbatim}
\usepackage{diagbox}
\usepackage{cleveref}
\usepackage{todonotes}
\usepackage{soul}
\usepackage[multiple]{footmisc}
\usepackage{multirow}
\usepackage{threeparttable}

\graphicspath{{./figures/}}

\DeclareCaptionSubType*[Alph]{table}
\DeclareCaptionLabelFormat{mystyle}{Table~\bothIfFirst{#1}{ ̃}#2}
\def\code#1{\texttt{#1}}
\DeclareMathOperator*{\argmin}{arg\!\min}

\title{
Incrementally Stochastic and Accelerated Gradient Information mixed Optimization for Manipulator Motion Planning
}

\author{Yichang Feng, Jin Wang$^{*}$, Haiyun Zhang, Guodong Lu 
\thanks{Manuscript received March 24, 2022; Accepted June 26, 2022. This paper was recommended for publication by Editor Kurniawati, Hanna upon evaluation of the Associate Editor and Reviewers’ comments. This work was supported by the Key R\&D Program of Zhejiang Province (2020C01025, 2020C01026), the National Natural Science Foundation of China (52175032), and Robotics Institute of Zhejiang University Grant (K11808). }
\thanks{The authors are with 1) State Key Laboratory of Fluid Power and Mechatronic Systems, School of Mechanical Engineering,
Zhejiang University, Hangzhou 310027, China;  2) Engineering Research Center for Design Engineering and Digital Twin of Zhejiang Province, School of Mechanical Engineering,
Zhejiang University, Hangzhou 310027, China. }
\thanks{$^*$Jin Wang, corresponding author, {\tt\small dwjcom@zju.edu.cn}}%
\thanks{Digital Object Identifier (DOI): see top of this page.}
}

\begin{document}

\markboth{IEEE Robotics and Automation Letters. Preprint Version. Accepted June, 2022}
{Feng \MakeLowercase{\textit{et al.}}: Incrementally Stochastic and Accelerated Gradient
Information mixed Optimization} 

%

\maketitle
\IEEEpeerreviewmaketitle

\begin{abstract}

This paper introduces a novel motion planner, incrementally stochastic and accelerated gradient information mixed optimization (iSAGO), for robotic manipulators in a narrow workspace. Primarily, we propose the overall scheme of iSAGO informed by the mixed momenta for an efficient constrained optimization based on the penalty method. In the stochastic part, we generate the adaptive stochastic momenta via the random selection of sub-functionals based on the adaptive momentum (Adam) method to solve the body-obstacle stuck case. Due to the slow convergence of the stochastic part, we integrate the accelerated gradient descent (AGD) to improve the planning efficiency. Moreover, we adopt the Bayesian tree inference (BTI) to transform the whole trajectory optimization (SAGO) into an incremental sub-trajectory optimization (iSAGO), which improves the computation efficiency and success rate further. Finally, we tune the key parameters and benchmark iSAGO against the other 5 planners on LBR-iiwa on a bookshelf and AUBO-i5 on a storage shelf. The result shows the highest success rate and moderate solving efficiency of iSAGO. 

\end{abstract}


\section{INTRODUCTION}

\IEEEPARstart{T}{he} industrial manufactory has widely applied robots in various areas such as welding and product loading or placing. Most of them always execute the above tasks under manual teaching or programming. So automatic production urges an efficient and robust optimal motion planning (OMP) algorithm to elevate industrial intelligence.  

Though former OMP studies gain a collision-free optimal trajectory for a safe and smooth motion by numerical optimization~\cite{Zucker2013CHOMP, Mukadam2018GPMP, Schulman2013SCO, Bhardwaj2020dGPMP} or probabilistic sampling~\cite{Kavraki1998PRMs, LaValle1998RRTs, Kuffner2000RRT-connect,Kalakrishnan2011STOMP, Karaman2011RRT*-PRM*, Rajendran2019CODES}, there still exist two main concerns this paper aims to solve: 

(i) \ul{Reliability}: The numerical methods, such as CHOMP\cite{Zucker2013CHOMP}, GPMP\cite{Mukadam2018GPMP}, and TrajOpt\cite{Schulman2013SCO}, can rapidly converge to a minimum with descent steps informed by the deterministic momenta/gradients. However, the local minima (i.e., failure plannings) 
are unavoidable with an inappropriate initial point. That is because the momenta information only describes the manifold of a local space near the initial point. So it is difficult for them to plan a safe motion with a high success rate (i.e., high reliability) in a narrow space. 

(ii) \ul{Efficiency}: The sampling method like STOMP\cite{Kalakrishnan2011STOMP} directly samples the trajectories to gain the optima. Others like RRT-Connect~\cite{Kuffner2000RRT-connect} grow a searching tree by the randomly sampled waypoints. Their sampling and wiring process can generate safe trajectories free of manifold information. However, their efficiency highly depends on the proportion the feasible subspace takes of the overall searching space. So the indiscriminate process takes enormous computation resources (i.e., low efficiency) in a narrow space. 

\begin{figure}[htb]
\begin{centering}
{\includegraphics[width=1\columnwidth]{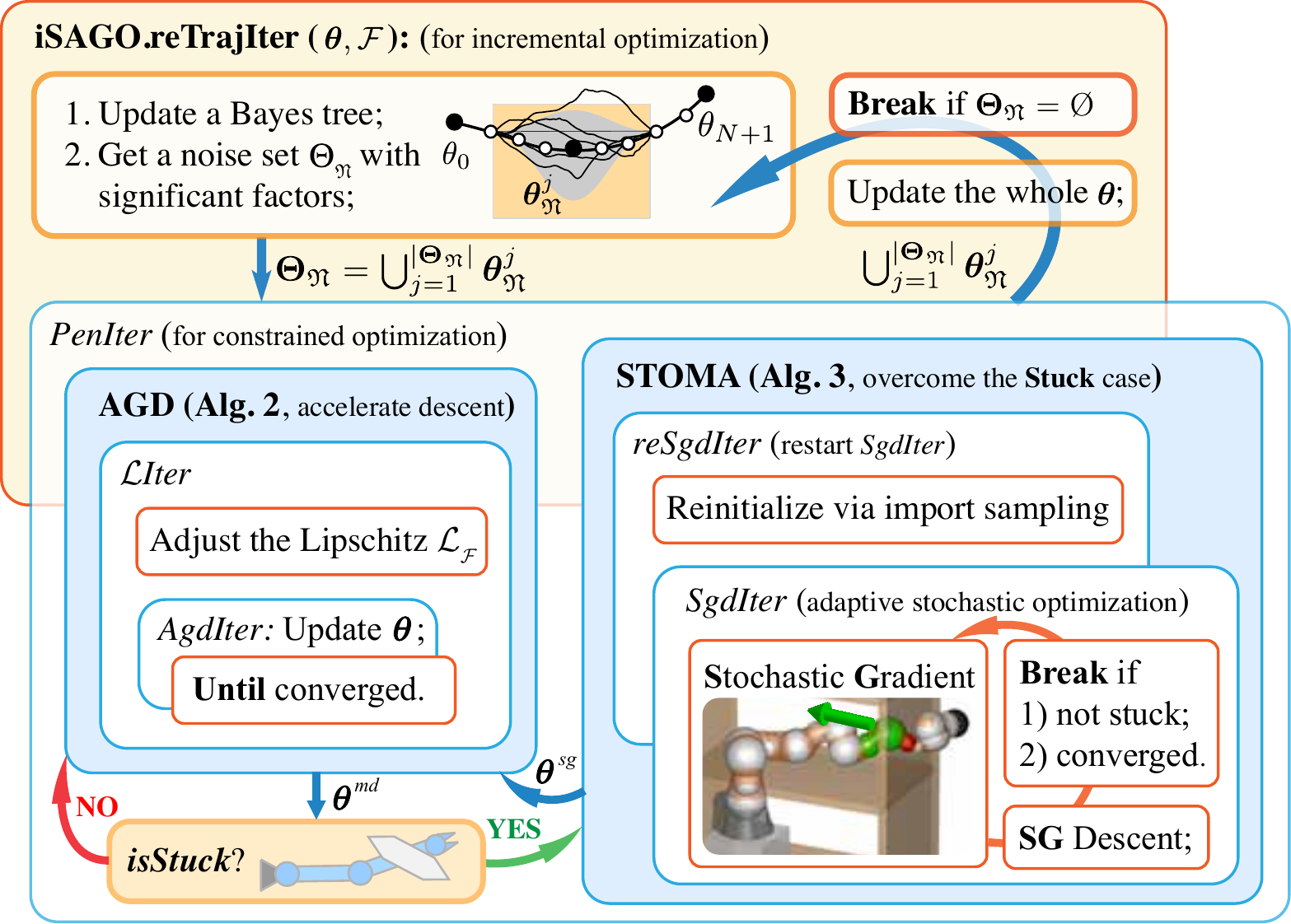}}
\par\end{centering}
\protect\caption{
A block diagram illustrates iSAGO (Algorithm~\ref{alg:iSAGO}, Section~\ref{sec:iSAGO}) that plans a collision-free smooth motion, whose outer layer shows how to refine a whole trajectory via BTI-based sub-trajectories optimization. The \textit{PenIter} integrates AGD (Algorithm~\ref{alg:L-reAGD}, Section~\ref{sec:L-reAGD}) and STOMA (Algorithm~\ref{alg:STOMA}, Section~\ref{sec:STOMA}) to solve the non-convex OMP by detecting the stuck case.
\label{fig:iSAGO}}
\end{figure}

This paper proposes \textbf{iSAGO} (Figure~\ref{fig:iSAGO}), which integrates the \textbf{S}tochastic and \textbf{A}ccelerated information of \textbf{G}radients and builds a Bayes tree for an \textbf{i}ncremental \textbf{O}ptimization:   
	
	(i) To address \ul{reliability}, \textbf{Stochastic trajectory optimization with moment adaptation (STOMA, \textnormal{Section~\ref{sec:STOMA}})} overcomes the local minima from the body-obstacle stuck cases by randomly selecting variables, such as collision-check balls, time intervals, and penalty factors. The information leakage of the stochastic momenta can somewhat modify OMP's manifold with fewer local minima.  
	
	(ii) Considering the low \ul{efficiency} of STOMA with $\mathcal{O}(\log N/N)$ convergence rate, iSAGO integrates \textbf{accelerated gradient descent (AGD, \textnormal{Section~\ref{sec:L-reAGD}})} in the non-stuck case because its $\mathcal{O}(1/N^{\frac{3}{2}})$ convergence rate of gradient norm is proven optimal in the first-order convex optimization. Furthermore, iSAGO adopts Bayes tree inference to \textbf{optimize the trajectory incrementally} (Section~\ref{sec:iSAGO}) for the further \ul{efficiency} elevation, which optimizes the convex or non-convex sub-trajectories separately in \textit{iSAGO.reTrajOpt}. 

The experiment of the 15 planning tasks with 44 problems on LBR-iiwa and AUBO-i5 (Section~\ref{sec:Evaluation}) validates higher reliability of iSAGO than the numerical methods~\cite{Zucker2013CHOMP, Mukadam2018GPMP, Schulman2013SCO} and its higher efficiency than the sampling methods~\cite{Kalakrishnan2011STOMP, Kuffner2000RRT-connect}.

\section{RELATED WORKS}

\subsection{Numerical Optimization}
The main concern of numerical optimization is rapidly descending to an optimum.  CHOMP~\cite{Zucker2013CHOMP}, ITOMP~\cite{Park2012ITOMP}, and GPMP~\cite{Mukadam2018GPMP} adopt the gradient descent method with the fixed step size. CHOMP uses Hamiltonian Monte Carlo (HMC)~\cite{Shirley2011HMC} for success rate improvement. To lower the computational cost, GPMP and dGPMP~\cite{Bhardwaj2020dGPMP} adopt iSAM2~\cite{Kaess2012iSAM2} to do incremental planning, and each sub-planning converges with a super-linear rate via Levenberg–Marquardt (LM)~\cite{Levenberg1944LM} algorithm. Meanwhile, ITOMP interleaves planning with task execution in a short-time period to adapt to the dynamic environment. Moreover, TrajOpt~\cite{Schulman2013SCO} uses the trust-region~\cite{Byrd2000TrustRegion} method to improve efficiency. It is also adopted by GOMP~\cite{Ichnowski2020GOMP} for grasp-optimized motion planning with multiple warm restarts learned from a deep neural network. Instead of deep learning, ISIMP~\cite{Kuntz2020ISIMP} interleaves sampling and interior-point optimization for planning. Nevertheless, the above methods may converge to local minima when the objective function is not strongly convex. 

AGD~\cite{Nesterov1983AG, Ghadimi2016AG-NLP} has recently been developed and implemented on the optimal control~\cite{Su2014ODE-AG, Wilson2018Lyapunov-AG} of a dynamic system described by differential equations whenever the objective is strongly convex or not. Moreover, stochastic optimization employs the momentum theorems of AGD and solves the large-scale semi-convex cases~\cite{Bottou2010SGD}, which generates stochastic sub-gradient via the random sub-datasets selection.  Adam~\cite{Kingma2014Adam} upgrades RMSProp~\cite{Hinton2012RMSProp} and introduces an exponential moving average (EMA) for momentum adaptation. Furthermore,  \cite{Wu2020Adam-UAV} adopts Adam to overcome the disturbance of a UAV system. 

\subsection{Probabilistic Sampling}
Unlike the numerical method, the sampling method constructs a search graph to query feasible solutions or iteratively samples trajectories for motion planning. 

PRM~\cite{Kavraki1998PRMs} and its asymptotically-optimal variants like PRM*~\cite{Karaman2011RRT*-PRM*} and RGGs~\cite{Solovey2018RGGs} make a collision-free connection among the feasible vertexes to construct a roadmap. Then they construct an optimal trajectory via shortest path (SP) algorithms like Dijkstra~\cite{Dijkstra1959Dijkstra-Alg} and Chehov~\cite{Hofmann2015Chehov}, which store and query partial trajectories efficiently. Unlike PRM associated with SP, RRT~\cite{LaValle1998RRTs} and its asymptotically-optimal variants like RRT*~\cite{Karaman2011RRT*-PRM*} and CODEs~\cite{Rajendran2019CODES} find a feasible solution by growing rapidly-exploring random trees (RRTs).

STOMP~\cite{Kalakrishnan2011STOMP} resamples trajectory obeying Gaussian distribution renewed by the important samples. \cite{Petrovic2019HGP-STO} introduces the GPMP's function to improve the searching efficiency of STOMP. Moreover, SMTO~\cite{Osa2020SMTO} applies Monte-Carlo optimization for multi-solution and refines them numerically.

\section{PROBLEM FORMULATION}

\subsection{Objective functional}\label{sec:costs}
We adopt the probabilistic inference model of GPMP~\cite{Mukadam2018GPMP} to infer a collision-free optimal trajectory given an environment with obstacles, a manipulator's arm, start, and goal. 

\subsubsection{GP prior}\label{sec:GP}
Gaussian process (GP) describes a multi-variant Gaussian distribution of trajectory $\bm{\theta}\backsim \mathcal{GP}(\bm{\mu},\bm{\mathcal{K}})$. $\bm{\mu}_\mathbf{T} = [\mu_i^{\top}]^{\top}\big|_{i\in\mathbf{T}}$ and  $\bm{\mathcal{K}}_\mathbf{T} = [\mathcal{K}_{i,j}]\big|_{i,j\in\mathbf{T}}$ determine its expectation value and covariance matrix of trajectory $\bm{\theta}_\mathbf{T} = [\theta_i^\top]^\top\big|_{i\in\mathbf{T}}$, respectively, where a set $\mathbf{T} = \{t_1,\dots, t_2\}$ denotes a period of time from $t_1$ to $t_2$. Since the above distribution originates from Gauss–Markov model (GMM)\footnote{Section 4 of \cite{Mukadam2018GPMP} introduces GP-prior, GMM generated by a linear time-varying stochastic differential equation (LTV-SDE), and GP-interpolation utilized by Section 5 for up-sampling. }, informed by the start and goal under zero acceleration assumption, we define the smooth functional
\begin{equation}\label{eq:gp_prior}
	\mathcal{F}_{gp}(\bm{\theta}_\mathbf{T}) = \frac{1}{2} \left(\bm{\theta}_\mathbf{T} - \bm{\mu}_\mathbf{T}\right)^\top \bm{\mathcal{K}_\mathbf{T}}^{-1} \left(\bm{\theta}_\mathbf{T} - \bm{\mu}_\mathbf{T}\right). 
\end{equation}

\subsubsection{Collision avoidance}\label{sec:obs}
Since the collision-free trajectory means a safe motion or successful planning, we first utilize $\bm{x}(\mathcal{B}_i, t)$ to map from a state $\theta_t$ at time $t$ to the position state of a collision-check ball (CCB-$\mathcal{B}_i$) on the manipulator. Then we calculate the collision cost $c(\bm{x})$, which increases when the distance between $\mathcal{B}_i$ and its closest obstacle decreases, and define the obstacle functional
\begin{equation}\label{eq:obs0}
	\mathcal{F}_{obs}(\bm{\theta}_\mathbf{T}) = \sum_{t \in \mathbf{T}} \sum_{\mathcal{B}_i \in \bm{\mathcal{B}}} c\left[\bm{x}(\mathcal{B}_i, t) \right] \cdot \left\| \dot{\bm{x}}(\mathcal{B}_i, t) \right\|, 
\end{equation}
where $\bm{\mathcal{B}} = \{\mathcal{B}_1,\dots, \mathcal{B}_{|\bm{\mathcal{B}}|}\}$ consists of $|\bm{\mathcal{B}}|$ CCBs. 

(i) \ul{Stuck case and local minima}: The functional~\eqref{eq:obs0} consists of discrete CCBs for calculating the cost and gradient. So there may exist the sub-gradients with opposite directions in a narrow workspace. In this case, the gradient of \eqref{eq:obs0} will approach zero when a collision cost is still high. It is a so-called {body-obstacle stuck case} that results in the local minima. According to Section~\ref{sec:SG} and the objective functional~\eqref{eq:chomp_cost}, all local minima are unsafe trajectories because the non-convexity originates from the obstacle part~\eqref{eq:obs0} caused by the stuck case rather than the convex GP part~\eqref{eq:gp_prior}. 

(ii) \ul{Continuous-time safety}: Though the above form combines the collision cost $c(\bm{x})$ with the change rate $\dot{\bm{x}}$ of CCB state for the continuous-time collision avoidance between two adjacent waypoints, it still cannot ensure safety when the obstacle allocation is sparse. 
This combination at one single time $t$ cannot precisely represent the obstacle functional within a continuous-time interval measured by an infinite number of timestamps, including $t$. 
So the up-sampling~\cite{Mukadam2018GPMP} method is adopted for continuous-time safety. According to GMM, we get $\bm{\Phi}(t+1, t)$ transforming $\theta_t$ to $\theta_{t+1}$ and
\begin{equation}\label{eq:Lambda}
	\bm{\Lambda}_{t,\tau} = \bm{\Phi}(t + \tfrac{\tau}{n_{t}^\textit{ip}+1}, t) - \bm{\Psi}_{t,\tau} \bm{\Phi}(t+1, t), 
\end{equation}
\begin{equation} \label{eq:Psi}
	\bm{\Psi}_{t,\tau} = \mathbf{Q}_{t,\tau} \bm{\Phi}(t+1,t + \tfrac{\tau}{n_{t}^\textit{ip}+1})^\top \mathbf{Q}_{t,1}^{-1}.
\end{equation}
where $\mathbf{Q}_{t,\tau}$ transforms the GP-kernel from $t$ to $(t + \tfrac{\tau}{n_{t}^\textit{ip}+1})$ with $\tau\in\{1,\dots,n_{t}^\textit{ip}\}$, and $n_{t}^\textit{ip}$ is the number of intervals between two adjacent waypoints $\{\theta_t,\theta_{t+1}\}$. Then we utilize $\bm\Lambda^{}_{t} = [\bm\Lambda_{t,1}^\top  \hdots \bm\Lambda_{t,n_{t_{}}^{\textit{ip}}}^\top ]^\top$, $\bm\Psi_{t}^{} = [  \bm\Psi_{t,1}^\top \hdots \bm\Psi_{t,n_{t}^{\textit{ip}}}^\top]^\top$ to form the upsampling matrix of trajectory $\bm{\theta}_\mathbf{T}$ in the period $\mathbf{T}$: 
\begin{equation}\label{eq:up_M}
	\hspace{-2mm}{
	\mathbf M_{} = \left[ \begin{array}{cccccccc}\hspace{-2mm}
	\mathbf I & \mathbf 0 & \mathbf 0 & \dots & \dots & \dots & \mathbf 0 & \mathbf 0\\
	\bm \Lambda_{t_1}^{} & \bm \Psi_{t_1}^{} & \mathbf 0 & \dots & \dots & \dots & \mathbf 0 & \mathbf 0\\
	\vdots & \vdots & \ddots &&&& \vdots& \vdots \\
	\mathbf 0 & \mathbf 0 & \dots & \mathbf I & \mathbf 0 & \dots & \mathbf 0 & \mathbf 0\\
	\mathbf 0 & \mathbf 0 & \dots & \bm \Lambda_{t}^{} & \bm \Psi_{t}^{} & \dots & \mathbf 0 & \mathbf 0\\
	\mathbf 0 & \mathbf 0 & \dots &  \mathbf 0 & \mathbf I& \dots & \mathbf 0 & \mathbf 0\\
	\vdots &\vdots  &&  & & \ddots& \vdots& \vdots \\
	\mathbf 0 & \mathbf 0 & \dots &  \dots & \dots& \dots & \bm \Lambda_{t_2\text{-}1}^{} & \bm \Psi_{t_2\text{-}1}^{} \\
	\mathbf 0 & \mathbf 0 & \dots &  \dots & \dots& \dots & \mathbf 0 & \mathbf I \\
	\end{array}\hspace{-1mm} \right]} \hspace{-2mm}
\end{equation}
for an upsampled trajectory $\bm{\theta}^\textit{up}_\mathbf{T} = \mathbf{M}(\bm{\theta}_\mathbf{T} - \bm{\mu}_\mathbf{T}) + \bm{\mu}^\textit{up}_\mathbf{T}$. Section~\ref{sec:SG} will detail the generation of a stochastic gradient in the time scale by the random selection of $\{n^\textit{ip}_{t}|_{t\in\mathbf{T}}\}$. 

\subsubsection{Objective functional}\label{sec:obj}
This paper adopts the objective functional of \cite{Zucker2013CHOMP, Mukadam2018GPMP} to generate an optimal trajectory $\bm{\theta}_\mathbf{T}$ in high smoothness without collision: 
\begin{equation} \label{eq:chomp_cost}
	\mathcal{F}(\bm{\theta}_\mathbf{T}) = \varrho \mathcal{F}_{gp}(\bm{\theta}_\mathbf{T}) + \mathcal{F}_{obs}(\bm{\theta}_\mathbf{T}^\textit{up}), 
\end{equation}
where $\varrho$ trades off between the GP prior (smoothness) and obstacle (collision-free) functional. Here we recommend \cite{Zucker2013CHOMP, Mukadam2018GPMP} for the detailed derivation of its gradient.

\subsection{Bayes tree construction}\label{sec:BT}
The above has introduced the relation between the stuck case and local minima, how to ensure the continuous-time safety, and how to transform the requirements of collision-free and smoothness to the Lagrangian formed objective functional of $\bm{\theta}_\mathbf{T}$. According to the definition of $\mathbf{T}$, we can group all timestamps $\{1,\dots,N\}$ into it to optimize the whole trajectory with $N$ waypoints or group some of them $\{t_1,\dots,t_2\}$ for sub-trajectories. These sub-problems can be solved incrementally to elevate the efficiency of trajectory optimization, like iGPMP's incremental replanning~\cite{Mukadam2018GPMP}. 

The Bayes tree (BT) definition in \cite{Mukadam2018GPMP} tells us a single chain with conjugated waypoints constructs a BT with a set of nodes $\Theta = \{\theta_t|_{t=1 \dots N}\}$ and branches $\{[\theta_t,\theta_{t+1}]|_{t=0 \dots N}\}$. In this way, we utilize \eqref{eq:chomp_cost} to calculate a BT-factor
\begin{equation}\label{eq:BT_factor}
	\mathcal{F}_t^\text{bt}= \mathcal{F}(\bm{\theta}_\mathbf{T}) \text{ with } \mathbf{T} = \{t-1, t, t+1\}, 
\end{equation}
informed by a minimum sub-trajectory. Section \ref{sec:iSAGO} will detail how to utilize it for incremental optimization.

\section{METHODOLOGY}


\subsection{Incremental optimization with mixed steps}\label{sec:iSAGO}

Given the start $\theta_0$ and goal $\theta_{N+1}$, iSAGO (Algorithm~\ref{alg:iSAGO}, Figure~\ref{fig:iSAGO}) first interpolates the support waypoints between them via the linear interpolation to gain an initial trajectory. Then it finds a series of collision-free waypoints facing the two planning scenarios: (i) the convex case satisfying
\begin{equation}\label{eq:convexHull}
	\mathcal{C}_{\theta} = \{ \bm{\theta}^{'} \bigr| \tfrac{|\mathcal{F}(\bm{\theta}^{\prime})-\mathcal{F}(\bm{\theta})-\langle\bar\nabla \mathcal{F}(\bm{\theta}), \bm{\theta}^{\prime}-\bm{\theta}\rangle|}{\|\bm{\theta}^{\prime}-\bm{\theta}\|^{2}} \leq \tfrac{\mathcal{L}_{\mathcal{F}}}{2} \};  
\end{equation}
(ii) the non-convex case where the trajectory gets stuck in obstacles, causing the local minima. 

Given these two cases, iSAGO mixes the accelerated and stochastic gradient information.  AGD (Section~\ref{sec:L-reAGD}) solves case (i) with an optimal convergence informed by \ul{the first order accelerated gradient}. On the other hand, STOMA (Section~\ref{sec:STOMA}) utilizes \ul{the stochastic gradient} to drag the trajectory from case (ii) into a convex sub-space, i.e., case~(i). Moreover, iSAGO uses the penalty method~\cite{Nocedal2006NumericalOpt} to nest the above methods in \textit{PenIter} for constrained optimization. 
\begin{algorithm}[]
\caption{iSAGO }\label{alg:iSAGO}
\DontPrintSemicolon
\LinesNumbered
\SetKwInOut{Input}{Input}
\SetKwInOut{Output}{Output}
\SetKwFunction{Union}{Union}
\Input {initial $\bm{\theta}$, $\mathcal{GP}$, BT, and tradeoff $\varrho$.}
\Output {optimized trajectory $\bm{\theta}$.}

\For(\tcp*[h]{\scriptsize incremental OMP}){$ \textit{reTrajIter}: i = 1 \dots N_{uf} $}{
Get noisy set $\bm{\Theta}_{\mathfrak{N}} = \bigcup_{j = 1}^{|\bm{\Theta}_{\mathfrak{N}}|}\bm{\theta}_{\mathfrak{N}}^{j}$ via~\eqref{eq:uniformFilter}; \;
\lIf {$\bm{\Theta}_{\mathfrak{N}} = \emptyset$}{
\Return. 
}
\For(\tcp*[h]{\small optimize $\bm{\Theta}_\mathfrak{N}$}){$j = 1 \dots |\bm{\Theta}_{\mathfrak{N}}|$}{
Get sub-trajectory $\bm{\theta}_\mathbf{T} \leftarrow \bm{\theta}_{\mathfrak{N}}^{j}$ with $\mathbf{T} \leftarrow \mathbf{T}^{j}_{\mathfrak{N}} $; \;
\For(\tcp*[h]{\footnotesize penalty method}){$ \textit{PenIter}: 1 \dots N_{\varrho} $}{
Check the body-obstacle stuck case according to Section~\ref{sec:SG}; \;
\lIf {isStuck}{
$\bm{\theta}_\mathbf{T} \leftarrow \code{STOMA($\bm{\theta}_\mathbf{T}$,$\mathcal{GP(\bm{\mu}_\mathbf{T},\bm{\mathcal{K}}_\mathbf{T})}$,$\varrho$)}$;
}\lElse{
$\bm{\theta}_\mathbf{T} \leftarrow \code{AGD($\bm{\theta}_\mathbf{T}$,$\mathcal{GP(\bm{\mu}_\mathbf{T},\bm{\mathcal{K}}_\mathbf{T})}$,$\varrho$)}$;
}
\lIf {$\mathcal{F}_{obs}(\bm{\theta}) < \textit{obs}\text{tol}$}{
\textbf{break};  
}\lElse{
$\varrho \leftarrow \kappa_{\varrho} \cdot \varrho$; 
}
}
Update the BT-factors $\{\mathcal{F}^\text{bt}_t |_{t\in\textbf{T}}\}$; \;
}}
\end{algorithm}

Formally, our mixed method optimizes the whole trajectory to plan a safe motion. However, it is inefficient because the trajectory consists of collision-free and in-collision (no-stuck and in-stuck) parts, each requiring a different method. So we adopt iSAM2~\cite{Kaess2012iSAM2} and build a BT (Section~\ref{sec:BT}) with factors~\eqref{eq:BT_factor}. Since an optimal BT has no significant factor, incremental-SAGO (iSAGO) finds the significant factors based on the mean and standard deviation of all factors: 
\begin{align}
	\mu_{\mathcal{F}} = \sum_{t = 1 \dots N}w_{t}\mathcal{F}_{t}^\text{bt},~
	\mathcal{D}_{\mathcal{F}} = \sqrt{\sum_{t = 1 \dots N}w_{t}\left(\mathcal{F}_{t}^\text{bt} - \mu_{\mathcal{F}} \right)^2}, 
\end{align}
where $w_{t} = 1/N$ and $N$ is the number of waypoints. Next, iSAGO gains a set of waypoints with significant factors:
\begin{equation}\label{eq:uniformFilter}
 {\Theta}_{\mathfrak{N}} = \left\{\theta_{t} \bigr|~ |\mathcal{F}_{t}^\text{bt}-\mu_{\mathcal{F}}| > c_\eta \mathcal{D}_{\mathcal{F}}, t = 1\dots N \right\},  
\end{equation}
where a smaller $c_\eta$ filters out more factors. Then it divides the whole trajectory into slices, each consisting of the adjacent $\theta\in\Theta_{\mathfrak{N}}$. After inserting one $\theta\notin\Theta_{\mathfrak{N}}$ at the head and tail of each slice, we get  $\bm{\Theta}_{\mathfrak{N}}=\bigcup_{j = 1}^{|\bm{\Theta}_{\mathfrak{N}}|}\bm{\theta}_{\mathfrak{N}}^{j}$ and stamp each sub-trajectory $\bm{\theta}_{\mathfrak{N}}^{j}$ by $\mathbf{T}^{j}_{\mathfrak{N}} = \{t\big| \theta_{t} \in \bm{\theta}_{\mathfrak{N}}^{j} \}$. When all sub-trajectories are optimized in \textit{iSAGO.reTrajIter}, iSAGO will update the BT and $\bm{\Theta}_{\mathfrak{N}}$ for the incremental optimization.


\subsection{Accelerated Gradient Descent}\label{sec:L-reAGD}

This paper gathers the accelerated gradient information for optimization because it only requires the first order momenta to achieve an optimal convergence~\cite{Nesterov2013AGintro} with the Lipschitz continuous gradient \eqref{eq:convexHull}. So we adopt the descent rules of \cite{Ghadimi2016AG-NLP}:
\begin{equation}\label{eq:AGDupdate}
\alpha_{k} = \frac{2}{k+1},~ \beta_{k} = \frac{1}{2 \mathcal{L}_{\mathcal{F}}},~ \lambda_{k } = \frac{k \beta_{k}}{2},
\end{equation} 
then for $\forall n > 1$, we have\footnote{See more details in \cite{Ghadimi2016AG-NLP}: (a) Lemma 1 provides an analytic view of the damped descent of Ghadimi's AGD; (b) Theorem 1 provides AGD's convergence rate, and its proof validates the accelerated descent process. } 
\begin{equation}\label{eq:AGD_converge1}
\min_{k = 1 \dots n} \|\bar{\nabla}\mathcal{F}(\bm{\theta}^{md}_k) \|^2 \leq \frac{96 \mathcal{L}_{\mathcal{F}} \|\bm{\theta}^{*} - \bm{\theta}_{0} \|^2 }{n^2(n+1)},  
\end{equation}

Some former studies~\cite{Nesterov1983AG, Nesterov2013AGintro, Ghadimi2016AG-NLP} set a fixed Lipschitz constant $\mathcal{L}_\mathcal{F}$ for AGD. However, \eqref{eq:AGD_converge1} indicates a strong correlation between the super-linear convergence rate and $\mathcal{L}_\mathcal{F}$. Recent studies~\cite{Su2014ODE-AG, Donoghue2015AR-AG} introduce the restart schemes for speed-up when the problem is not globally convex. Algorithm~\ref{alg:L-reAGD} adopts the trust-region method~\cite{Byrd2000TrustRegion} to adjust $\mathcal{L}_\mathcal{F}$ with factor $\kappa_{\tiny{\mathcal{L}}}$ and restart AGD until the condition\footnote{This paper uses $\mathcal{F}^{md}_k$, $\bar\nabla\mathcal{F}^{md}_k$, $\mathcal{F}^{ag}_k$ and $\mathcal{F}_k$ to simplify $\mathcal{F}(\bm{\theta}^{md}_k)$, $\bar\nabla\mathcal{F}(\bm{\theta}^{md}_k)$, $\mathcal{F}(\bm{\theta}^{ag}_k)$ and $\mathcal{F}(\bm{\theta}_k)$, while $\mathcal{F}^\text{bt}_i$ for the BT-factor at point $\theta_i$. }
\begin{equation} \label{eq:cond1}
	|\mathcal{F}_k-\mathcal{F}_{k-1}-\langle\bar\nabla \mathcal{F}_{k}^{md}, \bm{\theta}_{k}-\bm{\theta}_{k-1}\rangle| \leq \tfrac{\mathcal{L}_{\mathcal{F}}}{2}\|\bm{\theta}_{k}-\bm{\theta}_{k-1}\|^{2}
\end{equation}
is satisfied. In this way, we arbitrarily initialize $\mathcal{L}_{\mathcal{F}}$ as $\updelta_{0} \| \bar{\nabla} {\mathcal{F}}(\bm{\theta}_{0}) \|$ and update it for a robust convergence. 

\begin{algorithm}[]
\caption{AGD}\label{alg:L-reAGD}
\DontPrintSemicolon
\LinesNumbered
\SetKwInOut{Input}{Input}
\SetKwInOut{Output}{Output}
\SetKwFunction{Union}{Union}
\Input {initial $\bm{\theta}_{0}$, penalty ${\varrho}$, and $\mathcal{GP}(\bm{\mu},\bm{\mathcal{K}})$.} 
\Output{optimized $\bm{\theta}_{}^{md}$. }
\textbf{Initialize:}  \small{$\bm{\theta}_0^{ag} = \bm{\theta}_0$, $\bar\nabla\mathcal{F}_1^{md} = \bar\nabla\mathcal{F}_0$, $\mathcal{L}_{\mathcal{F}} = \updelta_{0} \| \bar{\nabla} {\mathcal{F}}_0 \|$}\normalsize; \;
\For(\tcp*[h]{\small adjust $\mathcal{L}_\mathcal{F}$}){$\textit{$\mathcal{L}$Iter}: j = 1 \dots N_{\mathcal{L}}$}{
\For(\tcp*[h]{\small AGD}){$\textit{AgdIter}: k = 1\dots N_{ag}$}{
$\bm{\theta}_{k}^{md}=\left(1-\alpha_{k}\right) \bm{\theta}_{k-1}^{ag}+\alpha_{k} \bm{\theta}_{k-1}$; \;
 \lIf{$k \geq 2$}{
 $\bar\nabla\mathcal{F}_{k}^{md} \leftarrow \bar\nabla\mathcal{F}(\bm{\theta}_k^{md})$;
 }
 Update $\alpha_{k}$, $\beta_{k}$ and $\lambda_{k}$ by \eqref{eq:AGDupdate}; \;
 $\bm{\theta}_{k}=\bm{\theta}_{k-1}-\lambda_{k} \bar\nabla\mathcal{F}_{k}^{md}$; \;
 $\bm{\theta}_{k}^{ag}= \bm{\theta}_{k}^{md}-\beta_{k} \bar\nabla\mathcal{F}_{k}^{md}$; \;
\lIf {Converged~\eqref{eq:conv} \textbf{or} \textit{isStuck}}{
\Return.
}
\If {condition~\eqref{eq:cond1} is not satisfied}{
$\left\{\bm{\theta}^{md}_{0}, \bm{\theta}^{ag}_{0}, \bm{\theta}_{0} \right\} \leftarrow \bm{\theta}^{md}_{k-1}$; 
$\mathcal{L}_\mathcal{F} \leftarrow \kappa_{\tiny{\mathcal{L}}}\cdot\mathcal{L}_\mathcal{F}$; $\bar\nabla\mathcal{F}_1^{md} \leftarrow \bar\nabla\mathcal{F}_{k-1}^{md}$; 
\textbf{break}
}
}
}
\end{algorithm}

The AGD in Algorithm~\ref{alg:L-reAGD} executes until it converges 
\begin{equation} \label{eq:conv}
	|\mathcal{F}_k - \mathcal{F}_{k-1}| < \mathcal{F}\text{tol},~ \|\bm{\theta}_{k}-\bm{\theta}_{k-1}\| < \theta\text{tol}
\end{equation}
or detects the stuck case according to Section~\ref{sec:SG}.

\subsection{Stochastic Optimization with Momenta Adaptation}\label{sec:STOMA}

\begin{figure*}[htbp]
\begin{centering}
	\begin{subfigure}[b]{0.45\textwidth}
		\centering
		\includegraphics[width=1\linewidth]{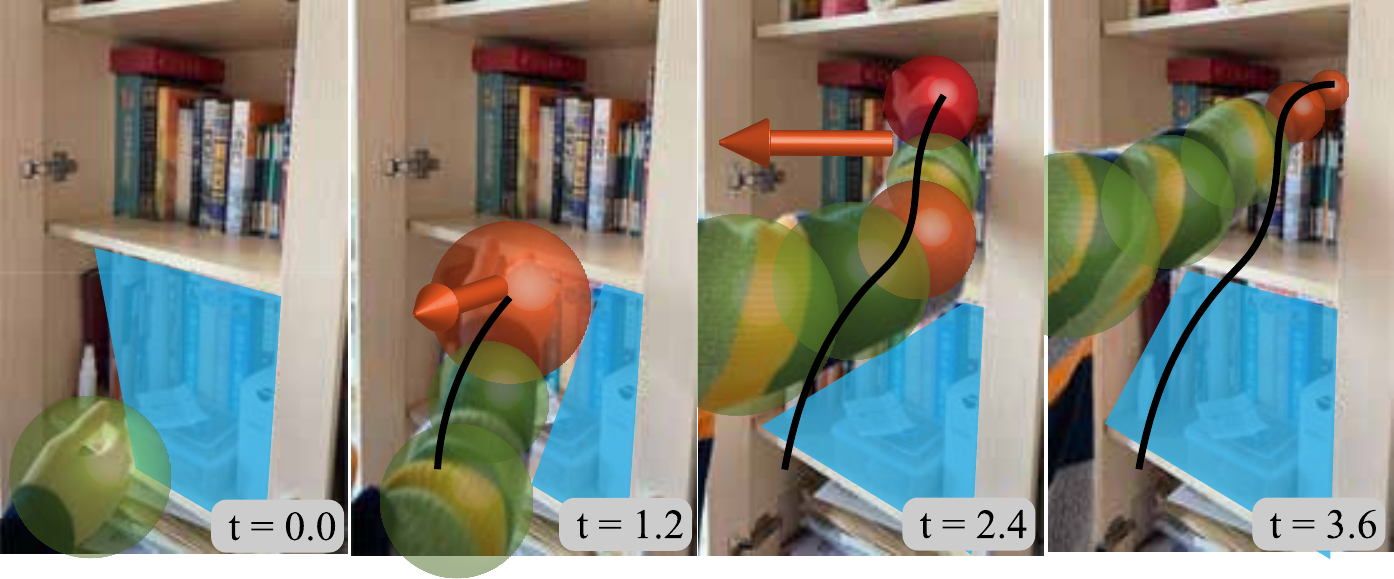}
		\caption{ A series of frames illustrate how the human arm reacts to obstacles when grasping inside a bookshelf. The green, orange, and red balls show none, low, and high-risk collision areas. 
		\label{fig:human_arm}}
	\end{subfigure}
	\hfill
	\begin{subfigure}[b]{0.50\textwidth}
		\centering
		\includegraphics[width=1\linewidth]{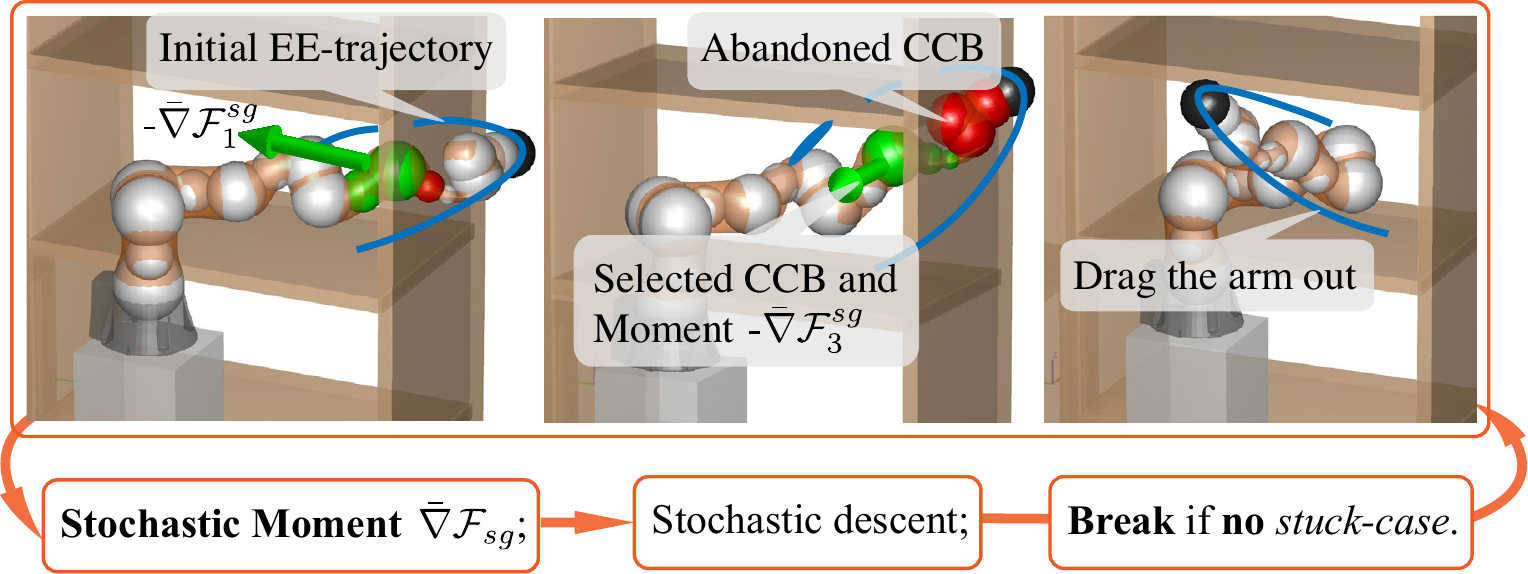}
		\caption{ STOMA drags the robot-arm out of the body-obstacle interference via adaptive stochastic descent. The green CCBs are selected for stochastic momenta (green arrow), while the red ones are abandoned.   
		\label{fig:STOMA}}
	\end{subfigure}
\end{centering}
\caption{The human-arm avoidance stimulated by the forces from the subparts inspires the robot-arm avoidance converged by stochastic gradient descent.  }
\end{figure*}

Some studies of deep learning~\cite{Duchi2011Ada,Hinton2012RMSProp,Kingma2014Adam,Goodfellow2016DL} propose stochastic gradient descent (SGD) for learning objects with noisy or sparse gradients. They optimize the objective function consisting of sub-functions in low coupling. Because of the higher robustness of Adam~\cite{Kingma2014Adam} during the function variation, we adopt its moment adaptation method and propose STOMA (Algorithm~\ref{alg:STOMA}, Figure~\ref{fig:STOMA}) in the stuck case. This section will first illustrate how to generate a stochastic gradient (SG), detect the stuck case, and introduce how to accelerate SGD with momentum adaptation. 
\begin{algorithm}[htp]
\caption{STOMA}\label{alg:STOMA}
\DontPrintSemicolon
\LinesNumbered
\SetKwInOut{Input}{Input}
\SetKwInOut{Output}{Output}
\SetKwFunction{Union}{Union}
\Input{initial $\bm{\theta}_0$,  $\varrho$ and $\mathcal{GP}(\bm{\mu}, \bm{\mathcal{K}})$. }
\Output{optimized $\bm{\theta}_{}^{sg}$. }

\For(\tcp*[h]{\small restart SGD}){\textit{reSgdIter}: $j = 1\dots N_{\textit{rsg}}$}{
 Sample $\bm{\Theta}_K = \bm{\theta}_0 \bigcup \{ \bm{\theta}_k \sim \mathcal{GP}\left(\bm{\mu}, \bm{\mathcal{K}}\right) \textbar_{ k = 1\dots K} \}$; \;
 Evaluate costs $\mathfrak{F}_K = \{ \mathcal{F} [\boldsymbol{\theta}_k]\big|_{k = 0,1,\dots,K}\}$; \;
 Initialize $k=0$, $\{\bm{\theta}_{0},\bm{\theta}_{0}^{ag}\} \leftarrow \argmin_{\bm{\theta} \in \bm{\Theta}_K} \mathfrak{F}_K$; \;
 \While(\tcp*[h]{\small adaptive SGD (\textit{SgdIter})}){True}{
 $k \leftarrow k+1$; Update $N_\textit{sg} \backsim \mathcal{U}(N_{lo}^\textit{sg}, N_{up}^\textit{sg})$; \;
 $\bm{\theta}_{k}^{sg}=\left(1-\alpha_{k}\right) \bm{\theta}_{k-1}^{ag}+\alpha_{k} \bm{\theta}_{k-1}$; \label{alg:STOMA:sg_step} \;
 Get $\bar\nabla\mathcal{F}^{sg}_{k}$ by \eqref{eq:SG}; check the \textbf{{Stuck}}-case; \;
\lIf {$\textbf{not}$ isStuck }{
 \Return. 
 }
 \If {$k \geq N_{\textit{sg}}$ \textbf{or} $\|\bm{\theta}_{k} - \bm{\theta}_{k-1}\| \leq \textit{SGtol}$}{
 Update $\bm{\theta}_0 = \bm{\theta}_{k}^{sg}$;  \textbf{break}
 }
 Estimate the $2^{nd}$ moment ${\bm{\mathfrak{M}}}_{k}$ via \eqref{eq:EMA}; \;
 Update $\alpha_{k}$, $\bm{\mathfrak{B}}_{k}$ and $\lambda_{k}$ via \eqref{eq:SGDupdate}; \;
 $\bm{\theta}_{k}=\bm{\theta}_{k-1}-\lambda_{k} \bar\nabla\mathcal{F}^{sg}_{k}$; \label{alg:STOMA:norm_step} \;
 $\bm{\theta}_{k}^{ag}= \bm{\theta}_{k}^{sg}-\bm{\mathfrak{B}}_{k} \bar\nabla\mathcal{F}^{sg}_{k}$; \label{alg:STOMA:ag_step} \;
 }
 }
\end{algorithm}
%

\subsubsection{SG generation and Stuck case check} \label{sec:SG}

Since \eqref{eq:chomp_cost} reveals that the whole objective functional uses the penalty factor $\varrho$ to gather the GP prior and obstacle information which accumulates the sub-functions calculated by collision-check balls (CCBs) in discrete time, we generate an SG  
\begin{equation} \label{eq:SG}
	\bar{\nabla}\mathcal{F}^{sg} = \hat{\varrho} \bar{\nabla}\mathcal{F}_{gp} +  \bar{\nabla}{\mathcal{F}_\textit{obs}^{sg}}
\end{equation}
in the {Functional}, {Time}, and {Space} scales:

(i) \ul{Functional}: We select the penalty factor $\frac{1}{\hat{\varrho}} \backsim \mathcal{U}(0, \frac{1}{\varrho})$. 

(ii) \ul{Time}: We select a set $\{n_{t}^\textit{ip} \backsim \mathcal{U}(0, N^{\textit{ip}})\big|_{t\in\mathbf{T}}\}$ whose element is the number of intervals of two support waypoints $[\theta_{t}^\top, \theta_{t+1}^\top]^\top$ to gain a sub-gradient for collision-check: 
\begin{equation}\label{eq:iobs_cost}
	\bar{\nabla}{\mathcal{F}_\textit{obs}^{sg}} = \mathbf M^\top\cdot \mathbf g_{up}, 
\end{equation}
where \eqref{eq:up_M} generates the upsampling matrix $\mathbf M$ with the set $\{n_{t}^\textit{ip} \big|_{t\in\mathbf{T}}\}$ to map the randomly upsampled gradient   
\begin{equation}
\nonumber
\small
\hspace{-1mm}
	\mathbf g_{up} = \left [
	\begin{array}{cccccc}
		\nabla^{\top} \mathcal{F}_{\textit{obs}}^{\bm{\mathcal{B}}}\left[{\theta}({t_1})\right], \\
		\nabla^{\top}\mathcal{F}_{\textit{obs}}^{\bm{\mathcal{B}}}\left[{\theta}(t_1+\frac{1}{n_{t_1}^{\textit{ip}}+1})\right], \dots,  \nabla^{\top}\mathcal{F}_{\textit{obs}}^{\bm{\mathcal{B}}} \left[{\theta}({t_1}+1)\right], \\
	 	\cdots \\
		\nabla^{\top}\mathcal{F}_{\textit{obs}}^{\bm{\mathcal{B}}}\left[{\theta}({t_2-\frac{n_{t_2\text{-}1}^{\textit{ip}}}{n_{t_2\text{-}1}^{\textit{ip}}+1}})\right], \dots,  \nabla^{\top} \mathcal{F}_{\textit{obs}}^{\bm{\mathcal{B}}} \left[{\theta}(t_2)\right] 
	 \end{array}
	 \hspace{-1mm} \right]^{\top}
\end{equation}
into the gradient $\bar{\nabla}\mathcal{F}_{\textit{obs}}^{sg}$ of $\bm{\theta}_\mathbf{T}$ with $\mathbf{T} = \{t_1,\dots,t_2\}$. 
	
(iii) \ul{Space}: The intuition of the Space rule comes from the collision avoidance of a human arm on a bookshelf in Figure~\ref{fig:human_arm}. When a participant stretches for an object, he makes sequential reflexes, stimulated by the exterior forces (orange arrow) affecting the danger parts (orange/red balls), in response to the body-obstacle collision. It indicates that the collision risk rises from shoulder to hand because the number of orange/red balls rises from shoulder to hand. So we first reconstruct each rigid body of a tandem manipulator by the geometrically connected CCBs. Next, we build each sub-problem (i.e., the collision avoidance of a single CCB) from shoulder to end-effector, considering the risk variation. Then ${\nabla}\mathcal{F}^{\bm{\mathcal{B}}}_{\textit{obs}}$ is calculated by accumulating the CCB-gradients from $\mathcal{B}_1$ nearby shoulder up to $\mathcal{B}_{k}$ nearby end-effector: 
\begin{equation}
\label{eq:SGD}
\begin{array}{c}
	\nabla \mathcal{F}_{obs}^{\bm{\mathcal{B}}}\left({\theta}_{t}\right) = \nabla \mathcal{F}_{obs}^{|\bm{\mathcal{B}}|}\left({\theta}_{t}\right), \\[8pt]
	\nabla \mathcal{F}_{obs}^{k} = \sum_{\mathcal{B}_i \in \bm{\mathcal{B}}_{k} \atop \phi_{i} \leq \phi_\text{tol} } \nabla_{\bm{q}_{i}} \mathcal{F}_{obs}, ~\bm{\mathcal{B}}_{k} = \{\mathcal{B}_1, \dots, \mathcal{B}_k \},  
\end{array}
\end{equation}
where $\bm{q}_{i}$ contains the joints actuating CCB-$\mathcal{B}_{i}$, and $\phi_{i}$ is the included angle between $\mathcal{B}_i$'s gradient ${\nabla}_{\bm{q}_{i}} \mathcal{F}_{obs}$ and the accumulated gradient ${\nabla}\mathcal{F}_{obs}^{i-1}$ (from $\mathcal{B}_1$ to $\mathcal{B}_{i-1}$). The random tolerance angle $\phi_\text{tol}$ generates SG in the Space scale by the random rejection of $\nabla_{\bm{q}_i} \mathcal{F}_{obs}$ with $\phi_i > \phi_\text{tol}$, meaning that the space stochasticity will vanish when $\phi_\text{tol} \rightarrow 180\degree$. 

\begin{figure}[htb]
\begin{centering}
{\includegraphics[width=1\columnwidth]{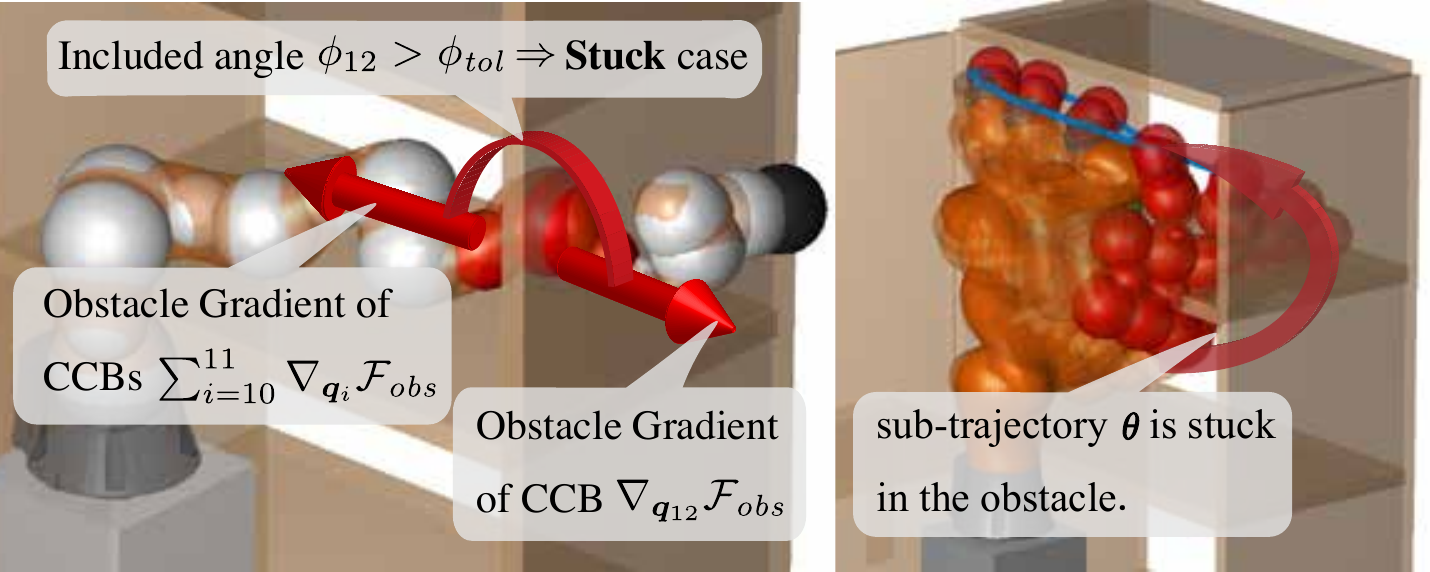}}
\par\end{centering}
\protect\caption{
The grey CCBs assemble an LBR-iiwa, while the red denotes an in-collision case. The stuck case occurs when the included angle $\phi > \phi_\text{tol}$.
\label{fig:stuckCase}}
\end{figure}

(iv) \ul{Stuck case}: Though, unlike the robot-arm, dragged out by the SG (Figure~\ref{fig:STOMA}), the stuck case cannot occur during the human-arm avoidance (Figure~\ref{fig:human_arm}), we can still imagine how confused a human-arm reflex is when a pair of opposite forces stimulate it. So we select a constant $\phi_\text{tol}$ to detect the stuck case confusing the collision-free planning.  Figure~\ref{fig:stuckCase} visualizes how an arm is stuck in an obstacle when $\exists \phi_i > \phi_\text{tol}$ and how a sub-trajectory containing a series of arms is stuck. We do not adopt the pair-wise check because the risk variates among different parts, and it is more efficient for the stuck check by a constant $\phi_\text{tol}$ to synchronize with the SG generation by a random $\phi_\text{tol}$. 
%

\subsubsection{Update rules}\label{sec:UpdateRule}

Adam~\cite{Kingma2014Adam} adopts the piecewise production of gradients and uses EMA from the initial step up to the current step and the bias correction method for the adaptive momentum estimation. Our work uses the squared $\ell_2$-norm of SG to estimate $\hat{\bm{\mathfrak{M}}}_1 = \|\bar{\nabla} \mathcal{F}^{sg}_1\|^2$ and adopt EMA and bias correction for 2nd-momenta $\{\bm{\mathfrak{M}}_{k}\}$ estimation: 
\begin{equation}\label{eq:EMA}
\hspace{-0mm}
\hat{\bm{\mathfrak{M}}}_{k} = \gamma \hat{\bm{\mathfrak{M}}}_{k-1} + (1-\gamma) \bar{\nabla} {\mathcal{F}^{sg}_{k}}^{\odot 2},~ \bm{\mathfrak{M}}_{k} = \frac{\hat{\bm{\mathfrak{M}}}_{k}} {1-\gamma^{k}}. 
\end{equation}

As shown in Algorithm~\ref{alg:STOMA}, we adopt the AGD rules~\cite{Ghadimi2016AG-NLP} rather than the EMA with bias correction for first-momentum adaption. $\{\alpha_{k}, \bm{\mathfrak{B}}_{k}, \lambda_{k}\}$'s update rules are also reset as 
\begin{equation}\label{eq:SGDupdate}
\alpha_{k} = \frac{2}{k+1},~ \bm{\mathfrak{B}}_{k} = \frac{\bm{\mathfrak{M}}_{k}^{-\frac{1}{2}}}{2 \delta},~ \frac{\lambda_{k}}{|\bm{\mathfrak{B}}_{k}|} \in\left[1,1+\frac{\alpha_{k}}{4} \right]. 
\end{equation}

In this way, the descent process, approximately bounded by a $\delta$-sized trust region, accelerates via the $\alpha$-linear interpolation between the conservative step and the accelerated step driven by $\{\lambda_{k}\}$ and $\{\bm{\mathfrak{B}}_{k}\}$ correspondingly. Then the process varies from the lag state guided by $\bm{\theta}$ to the shifting one guided by $\bm{\theta}^{ag}$ with $\{\alpha_{k}\}$ value. 
Since the SGD is roughly Lipschitz continuous, we get 
\begin{equation}\nonumber
\begin{aligned}
\mathcal{F}_{k-1} \leq & \mathcal{F}_{k} + \left\langle\bar{\nabla}\mathcal{F}_{k-1}, \bm{\theta}_{k} - \bm{\theta}_{k-1}\right\rangle\\
\leq& \mathcal{F}_{k} + \lambda_{k} \| \bar\nabla\mathcal{F}_{k}^{sg}\|^{2} + \lambda_{k}\left\|\Delta_{k}\right\| \left\|\bar\nabla\mathcal{F}_{k}^{sg}\right\|, \\
\end{aligned}
\end{equation}
%
where $\mathcal{F}_{k}$ and $\mathcal{F}_{k}^{sg}$ simplify $\mathcal{F}(\bm{\theta}_{k})$ and $\mathcal{F}(\bm{\theta}_{k}^{sg})$, and
\begin{equation}\nonumber
\begin{aligned}
	& \left\|\Delta_{k}\right\| = \| \bar\nabla\mathcal{F}_{k-1}-\bar\nabla \mathcal{F}_{k}^{sg} \| \leq \mathcal{L}_{\mathcal{F}}^{sg}(1-\alpha_{k}) \|\bm{\theta}_{k-1}^{ag}-\bm{\theta}_{k-1}\|, \\
	& \bm{\theta}_{k}^{ag}-\bm{\theta}_{k} = \left(1-\alpha_{k}\right)\left(\bm{\theta}_{k-1}^{ag}-\bm{\theta}_{k-1}\right)+\left(\lambda_{k}-\bm{\mathfrak{B}}_{k}\right) \bar\nabla\mathcal{F}_{k}^{sg}
\end{aligned}
\end{equation}
according to lines \ref{alg:STOMA:sg_step}, \ref{alg:STOMA:norm_step}, \ref{alg:STOMA:ag_step} of Algorithm~\ref{alg:STOMA}. Combining the Cauchy-Schwarz inequation, we get
\begin{equation} \nonumber 
	\begin{aligned}
		\mathcal{F}_{k-1}
		\leq & \mathcal{F}_{k} + \lambda_{k} \left(1 + \tfrac{\mathcal{L}_{\mathcal{F}}^{sg} \lambda_{k}}{2} \right) \left\|\bar\nabla \mathcal{F}_{k}^{sg}\right\|^{2} \\
		& + \tfrac{\mathcal{L}_{\mathcal{F}}^{sg}\left(1-\alpha_{k}\right)^{2}}{2} {\left\|\bm{\theta}_{k-1}^{ag} - \bm{\theta}_{k-1}\right\|^{2}}. 
	\end{aligned}
\end{equation}
Then through the accumulation from $k=1$ to $N$, we get 
\begin{equation}\label{ieq:SGD_converge01}
\nonumber 
\begin{array}{l}
	\frac{\mathcal{F}_{0} - \mathcal{F}_{N}}{\mathcal{L}_{\mathcal{F}}^{sg} }
	\leq \sum_{k=1}^{N}  \left(\lambda_{k} + \frac{\lambda_{k}^{2}}{2} +\frac{\left(\lambda_{k}-\beta_{k}\right)^{2}}{2 \Gamma_{k} \alpha_{k}} \sum_{l=k}^{N} \Gamma_{l} \right)
	\left\|\bar\nabla \mathcal{F}_{k}^{sg}\right\|^{2}
\end{array}  		
\end{equation}
where $\beta_{k} = |\bm{\mathfrak{B}}_{k}|_{\infty^{-}}$ and $\Gamma_{l} =  \left(1-\alpha_{l}\right) \Gamma_{l-1}$ with $\Gamma_0 = 1$. 

In sight of the stochasticity of $\mathcal{F}(\bm{\theta}^{sg})$ and its gradient $\bar{\nabla}\mathcal{F}^{sg}$ and presuming $\|\bar{\nabla}\mathcal{F}_{k}\| c_{l} \cdot \delta \geq \mathcal{L}_{\mathcal{F}}^{sg}$, we get 
\begin{equation}
	\frac{\mathcal{F}_{0} - \mathcal{F}_{N}}{N} 
	\leq \frac{5 \mathbf{G}_{\infty} }{4\delta} \left( 1 + \frac{11 c_{l}}{32} \right)  
	\frac{\ln (N+1)}{N}, 
\end{equation}
where $c_{l}$ is the fraction of $\mathcal{L}^{sg}_{\mathcal{F}}$ and trust region box $\delta \cdot \|\bar{\nabla}\mathcal{F}_{k}\|$, and $\mathbf{G}_{\infty} \leq \mathds{E}(\|[\bar{\nabla}^\top\mathcal{F}_{1}^{sg},\dots,\bar{\nabla}^\top \mathcal{F}_{N}^{sg}]\|_{\infty})$.  

Figure~\ref{fig:STOMA} illustrates how a robot reflexes to the stuck case like the human in Figure~\ref{fig:human_arm}, following the update rules in \textit{SgdIter}. To improve the efficiency and reliability, we nest the above SGD (i.e., \textit{SgdIter}) in \textit{reSgdIter}. Once the SGD is restarted, we select the trajectory with the lowest cost from the sample set $\bm{\Theta}_K$ to reinitialize the next SGD, whose maximum number $N_\textit{sg}$ is selected randomly in $[N_{lo}^\textit{sg}, N_{up}^\textit{sg}]$.

\section{EXPERIMENT}

This section will first introduce the objective functional implementation in Section~\ref{sec:Imp}. Then it will detail the benchmark and parameter setting in Sections~\ref{sec:setup}~\&~\ref{sec:parameters} and analyze the benchmark results in Section~\ref{sec:analysis}. 

\subsection{Implementation details}\label{sec:Imp}

\subsubsection{GP prior}\label{sec:const_gp_prior}
We assume the velocity is constant and gain the GMM and GP-prior \eqref{eq:gp_prior} to reduce the dimension. Our benchmark uses $N=12$ support states with $N^\textit{ip}=8$ intervals (116 states in total) for iSAGO and GPMP2, 116 states for STOMP and CHOMP, 12 and 58 states for TrajOpt\footnote{Since TrajOpt proposes a swept-out area between support waypoints for continuous-time safety, we choose TrajOpt-12 for safety validation and TrajOpt-58 to ensure the dimensional consistency of the benchmark. }, and \code{Max\-Connection\-Distance} $=\frac{\|\theta_{g} - \theta_{0}\|}{12}$ for RRT-Connect. All implementations except RRT-Connect apply linear interpolation rather than manual selection to initialize trajectory for a fair comparison. 

\subsubsection{Collision cost function}\label{sec:collisionCost}
We adopt the polynomial piecewise form of \cite{Zucker2013CHOMP} for the collision cost $c(\bm{x})$ of \eqref{eq:obs0}. 

\subsubsection{Motion constraints}
The motion cost function in GPMP~\cite{Mukadam2018GPMP} is adopted to drive the robot under the preplanned optimal collision-free trajectory in the real world. 

\subsection{Evaluation} \label{sec:Evaluation}

\subsubsection{Setup for benchmark}\label{sec:setup}
This paper benchmarks iSAGO against the numerical planners (CHOMP~\cite{Zucker2013CHOMP}, TrajOpt~\cite{Schulman2013SCO}, GPMP2~\cite{Mukadam2018GPMP}) and the sampling planners (STOMP~\cite{Kalakrishnan2011STOMP}, RRT-Connect~\cite{Kuffner2000RRT-connect}) on  a 7-DoF robot (LBR-iiwa)  and a 6-DoF robot (AUBO-i5). Since the benchmark executes in MATLAB, we use \code{BatchTrajOptimize3DArm} of GPMP2-toolbox to implement GPMP2, \code{plannarBiRRT} with 20s maximum time for RRT-Connect, \code{fmincon}\footnote{We use \code{optimoptions('fmincon','Algorithm','trust\--region\--reflective','Specify\-Objective\-Gradient', true)} to apply \code{fmincon} for the trust-region method like TrajOpt and calculates \code{Objective\-Gradient} analytically and \code{Aeq} (a matrix whose rows are the constraint gradients) by numerical differentiation.} for TrajOpt, \code{hmcSampler}\footnote{Our benchmark defines an \code{hmcSampler} object whose logarithm probabilistic density function \code{logpdf} is defined by \eqref{eq:chomp_cost}, uses \code{hmcSampler.drawSamples} for HMC adopted by CHOMP. } for CHOMP, and \code{mvnrnd}\footnote{STOMP samples the noise trajectories by \code{mvnrnd} and updates the trajectory via projected weighted averaging. } for STOMP. Since all of them are highly tuned in their own studies, our benchmark uses their default settings. 

\begin{figure}[h]
	\begin{centering}
		\begin{subfigure}[b]{0.23\textwidth}
			\centering
			\includegraphics[width=1\linewidth]{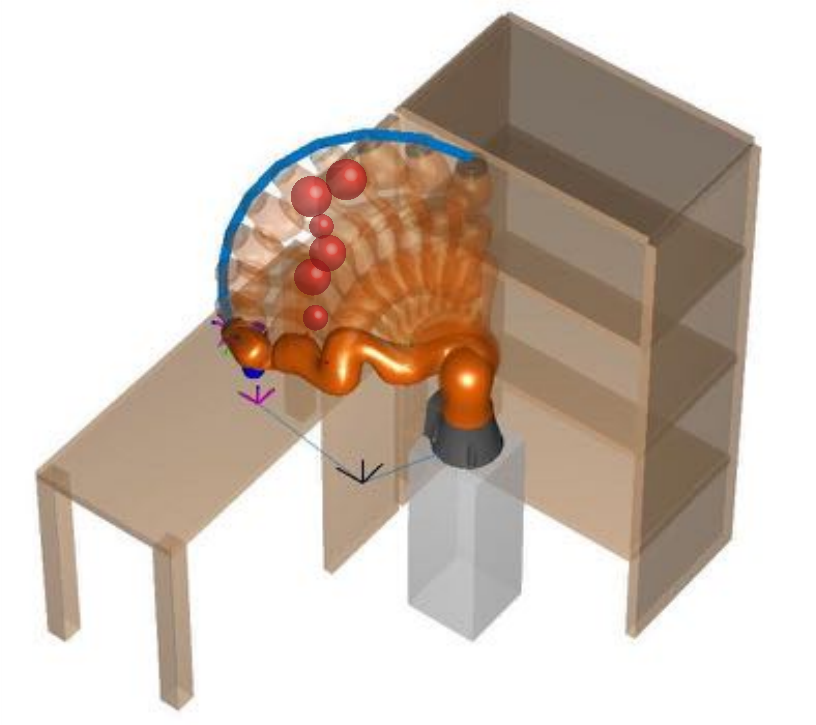}
			\caption{Task A-1, 6 red CCBs}
			\label{fig:iiwa:A1}
		\end{subfigure}
		\hfill
		\begin{subfigure}[b]{0.23\textwidth}
			\centering
			\includegraphics[width=1\linewidth]{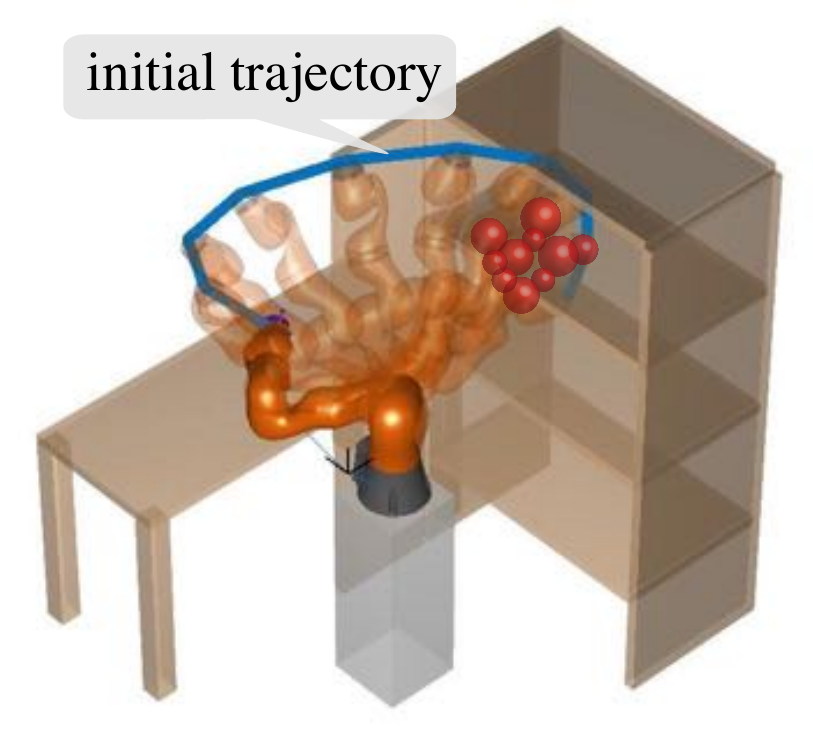}
			\caption{Task B-1, 11 red CCBs}
		\end{subfigure}
		\hfill
		\begin{subfigure}[b]{0.23\textwidth}
			\centering
			\includegraphics[width=1\linewidth]{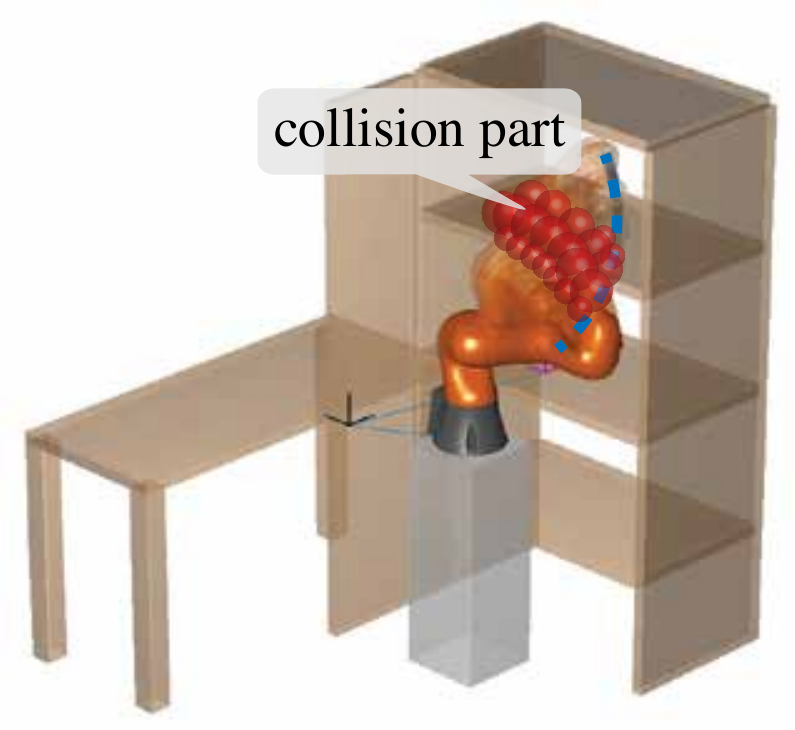}
			\caption{Task C-3, 22 red CCBs}
			\label{fig:iiwa:C3}
		\end{subfigure}
		\hfill
		\begin{subfigure}[b]{0.23\textwidth}
			\centering
			\includegraphics[width=1\linewidth]{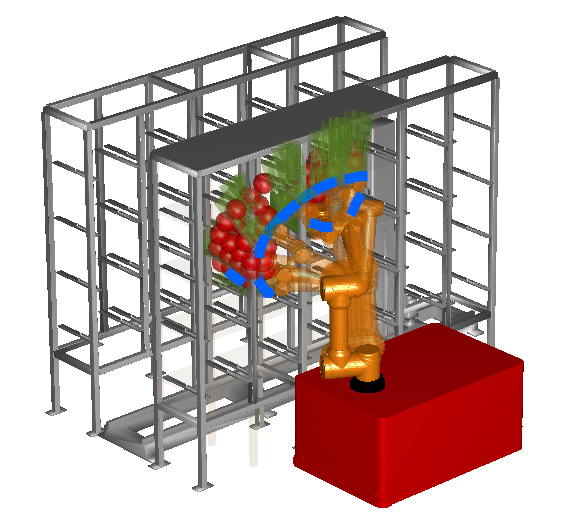}
			\caption{Task C-6, 35 red CCBs}
			\label{fig:AUBO:C6}
		\end{subfigure}
	\end{centering}
	\caption{The initial trajectory with red collision parts visualizes our benchmark on LBR-iiwa or AUBO-i5. \textit{Task C-6} means the No.{6} task of class {C}. 
	\label{fig:problems_ABC}}
\end{figure}

To illustrate the competence of iSAGO for planning tasks, we conduct 25 experiments on LBR-iiwa at a bookshelf and 12 experiments on AUBO-i5 at a storage shelf. We categorize all tasks into 3 classes (A, B, and C) whose planning difficulty rises with the stuck cases in the initial trajectory increase. Figure~\ref{fig:problems_ABC} visually validates our classification because the number of red in-collision CCBs increases from Task A-1 to Task C-3. Considering the difficulty of different classes, we first generate 2 tasks of class A, 3 tasks of class B, and 4 tasks of class C for LBR-iiwa (Figures~\ref{fig:iiwa:A1}-\ref{fig:iiwa:C3}). Then we generate 6 tasks of class C for AUBO-i5 (Figures~\ref{fig:AUBO:C6}). Each task consists of 2 to 4 problems\footnote{Since LBR-iiwa has 7 DoFs, meaning infinite solutions for the same goal constraint, our benchmark uses LM~\cite{Levenberg1944LM} with random initial points to generate different goal states. So one planning task of LBR-iiwa with the same goal constraint has several problems with different goal states. Meanwhile, each task of AUBO-i5 has two problems with the same initial state and goal constraint because the same goal constraint has only one solution for the 6-DoF, and we switch yaw-angle in $\{0\degree, 180\degree\}$ for each. } with the same initial state and goal constraint. Moreover, we compare iSAGO with SAGO to show the efficiency of the incremental method and compare SAGO with STOMA or AGD to show the efficiency or reliability of the mixed optimization. 
\begin{figure}[htb]
	\begin{centering}
		\begin{subfigure}[b]{0.24\textwidth}
			\centering
			\includegraphics[width=1\linewidth]{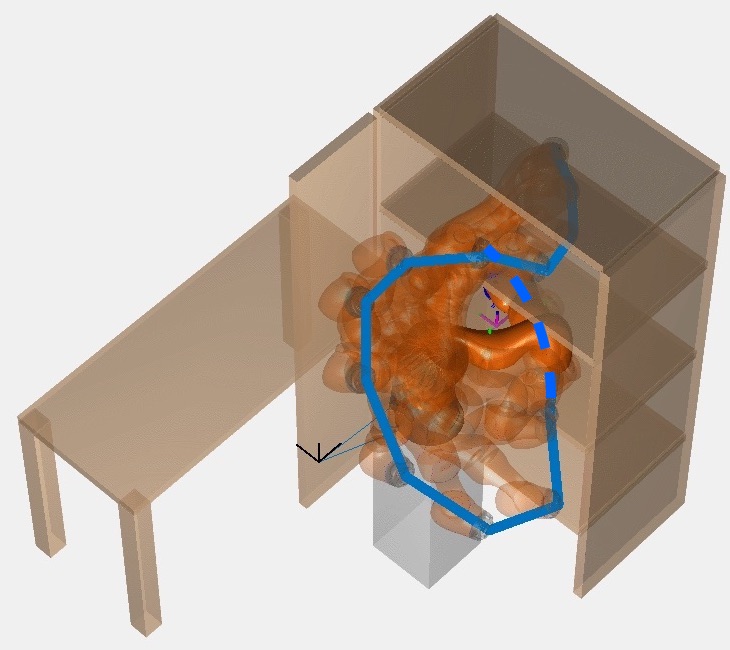}
			\caption{Trajectory C-3.2}
			\label{fig:iiwa:C3.2}
		\end{subfigure}
		\hfill
		\begin{subfigure}[b]{0.22\textwidth}
			\centering
			\includegraphics[width=1\linewidth]{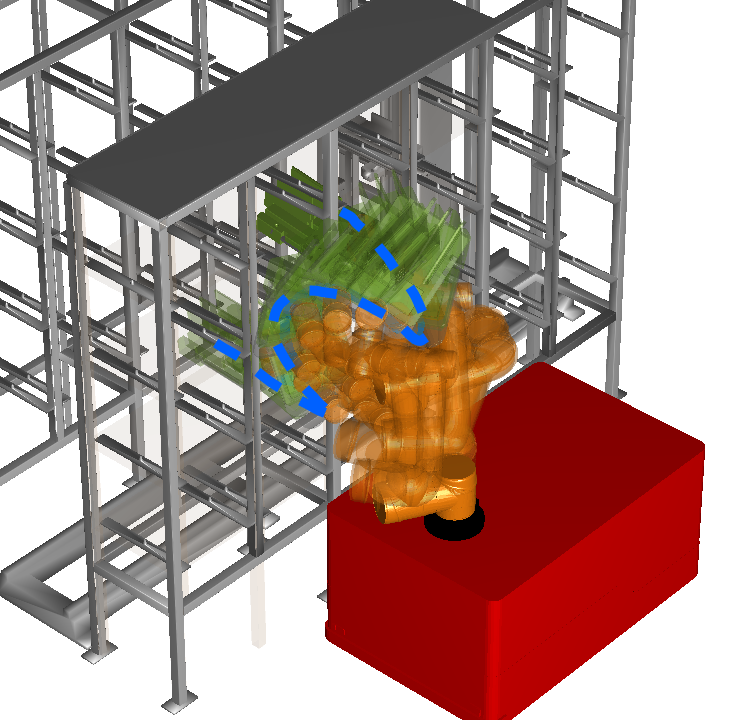}
			\caption{Trajectory C-6.2}
			\label{fig:AUBO:C6.2}
		\end{subfigure}	
		\hfill
	\end{centering}
	\caption{Some results of iSAGO in class C  (\textit{Trajectory C-3.2} means the 2nd planning result of task 3 in class C. )  
	\label{fig:results_C}}
\end{figure}
%

\subsubsection{Parameter setting}\label{sec:parameters}

Since STOMA (Algorithm~\ref{alg:STOMA}, Figure~\ref{fig:STOMA}) proceeds SGD roughly contained in a $\delta$-sized trust-region with $\gamma$-damped EMA of historical momenta, this section first tunes these two parameters, then the others.

According to abundant experiences, we tune the key parameters $\delta \in [0.04, 0.80]$, $\gamma \in [0.50, 0.99]$ and conduct 10 different class C tasks for each pair of parameters. Table~\ref{table:results_A} indicates that the SGD rate increases with $\delta$-expansion. In contrast, the success rate decreases with the excessive $\delta$ (i.e., too high or too low). Moreover, the SGD rate increases with $\gamma$-shrink while the success rate decreases with $\gamma$-shrink. All in all, iSAGO performs more stable with a smaller step-size $\delta$, hampering local minima's overcoming. Meanwhile, it combines less historical momenta to perform faster, reducing the reliability. So we set $\delta = 0.40$ and $\gamma = 0.90$.  
\begin{table}
\caption{Results for $\{\delta, \gamma\}$ tuning of 10 tasks in class \textbf{C}}
\label{table:results_A}
\centering
\begin{threeparttable}
{
\begin{tabular}{l|ccccccc}
\toprule
\scalebox{0.8}{\diagbox{$\gamma$}{(\%)|(s)}{$\delta$}} & 0.80 & 0.40 & 0.08 & 0.04 \\ 
\midrule
 0.50 & 80 | 1.253 & 90 | 1.898 & 60 | 4.198 & 40 | 6.309 \\
 0.90 & 90 | 2.512 & \bf{100} | \bf{2.463} & 80 | 4.973 & 60 | 8.219 \\
 0.99 & 90 | {2.672} & \bf{100} \textnormal{| 3.089} & 80 | 7.131 & 60 | 12.57 \\
\bottomrule
\end{tabular}}
\begin{tablenotes}\scriptsize
     \item[1] (\%) and (s) denote the success rate and average computation time. 
\end{tablenotes}
\end{threeparttable}
\end{table}

Besides $\delta$ and $\gamma$, we set the initial penalty $\varrho = 1.25\text{e-2}$, penalty factor $\kappa_{\varrho} = 0.4$, and filter factor $c_\eta = 2.0$ in Algorithm~\ref{alg:iSAGO}. 
In Algorithm~\ref{alg:L-reAGD}, we set the tolerances $\mathcal{F}\text{tol} = 8\text{e-4}$, $\theta \text{tol} = \text{1e-3}$, $\textit{obs}\text{tol} = \text{1e-4}$ and scaling factor $\kappa_{\mathcal{L}} = 6.67$. 
In Algorithm~\ref{alg:STOMA}, we set the number of samples $K = 12$, tolerance $\textit{SG}\text{tol} = 6.4\text{e-3}$, and the number of \textit{SgdIter} $N_\textit{sg} \backsim \mathcal{U}(35, 55)$. Moreover, we select $\phi_\text{tol} = 95 \degree$ to check stuck case while $\phi_\text{tol}\backsim\mathcal{U}(60\degree, 180 \degree)$ to generate SG.

\subsubsection{Result analysis}\label{sec:analysis}

\begin{table*}[htp]
\caption{Results of 15 tasks (44 planning problems) with 5 repeated tests}
\label{table:results_C:iiwa}
\centering
\begin{threeparttable}
\scalebox{1}{
\begin{tabular}{llcccc|cccc|cc}
\toprule
& \multirow{2}*{Problem} & \multicolumn{4}{c|}{Our incremental mixed optimization}  & \multicolumn{4}{c|}{numerical optimization} & \multicolumn{2}{c}{probabilistic sampling} \\
& & \bf{AGD} & \bf{STOMA} & \bf{SAGO} & \bf{iSAGO} & {TrajOpt-12} & {TrajOpt-58} & {GPMP2-12} & {CHOMP} & {STOMP} & {RRT-Connect} \\ 
\midrule
\multirow{3}*{Scr(\%)} 
& iiwa\_AB & 75 & 97.5 & \bf{100} & \bf{100} & 27.5  & 77.5 & 81.25 & 87.5 & {82.5} & \bf{100}\\
& iiwa\_C & 8.75 & 87.5 & 91.25 & \bf{93.75} & 3.75 &  8.75 & 11.25 & 30 & {25} & 72.5 \\
& aubo\_C & 5 & 71.67 & 83.33 & \bf{88.33} & 3.33 &  6.67 & 11.67 & 26.67 & {25} & 73.33 \\
\midrule
\multirow{3}*{Avt(s)} 
& iiwa\_AB & 0.278 & 3.031 & 1.714 & 1.145 & \bf{0.127} & {0.495} & 1.294 & 2.728 & 3.080 & 8.92 \\
& iiwa\_C & 0.288 & 7.622 & 4.284 & 2.132 & \bf{0.232} & {1.108}  & 2.464 & 6.989 & 7.691 & 13.65 \\
& aubo\_C & 0.343 & 8.584 & 4.812 & 2.413 & \bf{0.245} & {1.119}  & 2.610 & 7.588 & 8.640 & 16.02 \\
\midrule
\multirow{3}*{Sdt(s)} 
& iiwa\_AB & 0.032 & 0.823 & 0.362 & 0.287 & \bf{0.034} & {0.076} & 0.845 & 0.988 & 2.540 & 3.613 \\
& iiwa\_C & 0.045 & 1.735 & 1.048 &  0.319 & \bf{0.055} & 0.101 & 1.312 & 1.290 & 5.233 & 4.562 \\
& aubo\_C & 0.078 & 1.821 & 1.101 &  0.334 & \bf{0.057} & 0.106 & 1.377 & 1.354 & 5.494 & 3.974 \\
\bottomrule
\end{tabular}}
\begin{tablenotes}\scriptsize
     \item[1] Scr(\%), Avt(s) and Sdt(s) denote the success rate, average computation time and standard deviation of computation time, respectively. 
     \item[2] iiwa\_AB, iiwa\_C and aubo\_C denote 16 class A\&B problems on LBR-iiwa, 16 class C problems on LBR-iiwa, and 12 class C problems on AUBO-i5, respectively. 
\end{tablenotes}
\end{threeparttable}
\end{table*}

Figure~\ref{fig:iiwa:C3.2} shows how iSAGO drags a series of in-stuck arms of Figure~\ref{fig:iiwa:C3} out of the bookshelf by STOMA to grasp a cup located in the middle layer of the bookshelf safely. Meanwhile, Figure~\ref{fig:AUBO:C6.2} shows how AUBO-i5 avoids the collision of Figure~\ref{fig:AUBO:C6} to transfer the green piece between different cells of the storage shelf safely. 

Table~\ref{table:results_C:iiwa} shows iSAGO gains the highest success rate compared to the others. The random \textit{SgdIter} number $N_\textit{sg}$ and AGD help iSAGO gain the fourth solving rate after TrajOpt-12, TrajOpt-58, and GPMP2-12. Though TrajOpt-58 compensates for the continuous safety information leakage of TrajOpt-12 in iiwa\_AB, it still cannot escape from the local minima contained by the trust region and gains the second lowest success rate just above TrajOpt-12 in iiwa\_C/aubo\_C. In contrast, iSAGO successfully descends into an optimum with adaptive stochastic momenta and an appropriate initial trust-region. Thanks to HMC, randomly gaining the Hamiltonian momenta, CHOMP approaches the optimum with the third-highest success rate, whose failures are informed by the deterministic rather than the stochastic gradients. RRT-Connect with a limited time has the highest and the second-highest success rate in iiwa\_AB and iiwa\_C/aubo\_C, respectively. However, a higher rate needs a smaller connection distance which restricts the RRT growth and computation efficiency. Though STOMP is free of gradient calculation, the significant time it takes to resample does a minor effect on feasible searching, mainly limited by the Gauss kernel. As for GPMP2-12 and TrajOpt-58, the LM and trust-region methods help approach the stationary point rapidly. In contrast, the point of iiwa\_C/aubo\_C has significantly lower feasibility than that in iiwa\_A because the initial trajectory of iiwa\_C/aubo\_C gets stuck deeper. 

The comparison between STOMA and SAGO in Table~\ref{table:results_C:iiwa} shows how AGD performs a 45\% accelerated descent towards an optimum. The comparison between iSAGO and SAGO indicates a 55\% higher efficiency of incremental planning. The 90\% lower success rate of AGD than SAGO shows the limitation of the numerical method. It validates that STOMA can modify the manifold with less local minima by randomly selecting sub-functional. 
 
The comparison between iiwa\_C and aubo\_C in Tables~\ref{table:results_C:iiwa} shows that iSAGO performs better when the feasible solution is nearby the motion constraints. That is because the motion constraint somewhat stabilizes the $\delta$-sized SGD of STOMA. So the large $\delta$-sized steps bring a severe disturbance and weaken STOMA's performance by 20\% when obstacles wrap the feasible solution. Meanwhile, the comparison between iiwa\_AB and iiwa\_C shows that the stuck case reduces the success rate of the numerical method by 90\% and validates iSAGO's higher reliability in a narrow workspace under the same non-manual initial condition.

\section{CONCLUSIONS}

iSAGO utilizes the mixed information of accelerated (AGD) and stochastic (STOMA) momentum to overcome the body-obstacle stuck cases in a narrow workspace. 

(i) STOMA performs an adaptive stochastic descent to avoid the local minima confronted by the numerical methods, and the results show the highest success rate among them. 

(ii) AGD integrated by iSAGO accelerates the descent informed by the first-order momenta. The results show it saves the sampling method's computation resources and gain the fourth solving rate.

(iii) The incremental planning optimizes the sub-planning to elevate the whole-planning rate further. The results show the 55\% higher efficiency of iSAGO than SAGO.







\bibliographystyle{IEEEtran}
\bibliography{SAGO_bib}

\end{document}